\documentclass{article} 
 
\usepackage{iclr_fmwild,times}
\iclrfinaltrue  

\usepackage{amsmath,amsfonts,bm}

\def\eqref#1{equation~\ref{#1}}

\def\1{\bm{1}}

\DeclareMathAlphabet{\mathsfit}{\encodingdefault}{\sfdefault}{m}{sl}
\SetMathAlphabet{\mathsfit}{bold}{\encodingdefault}{\sfdefault}{bx}{n}

\usepackage{hyperref}
\usepackage{url}

\usepackage{booktabs}       
\usepackage{amsfonts}       
\usepackage{nicefrac}       
\usepackage{microtype}      
\usepackage{amsmath}
\usepackage{multirow}
\usepackage{xcolor}
\usepackage{graphicx}
\usepackage{wrapfig}
\usepackage{float}

\title{MetaSC: Test-Time Safety Specification \\ Optimization for Language Models}

\author{Víctor Gallego  \\
Komorebi AI. \emph{Madrid, Spain.} \\
\texttt{victor.gallego@komorebi.ai} \\
}

\begin{document}

\maketitle

\begin{abstract}
We propose a novel dynamic safety framework that optimizes language model (LM) safety reasoning at inference time without modifying model weights. Building on recent advances in self-critique methods, our approach leverages a meta-critique mechanism that iteratively updates safety prompts—termed specifications—to drive the critique and revision process adaptively. This test-time optimization not only improves performance against adversarial jailbreak requests but also in diverse general safety-related tasks, such as avoiding moral harm or pursuing honest responses. Our empirical evaluations across several language models demonstrate that dynamically optimized safety prompts yield significantly higher safety scores compared to fixed system prompts and static self-critique defenses. Code released at \href{https://github.com/vicgalle/meta-self-critique.git}{\texttt{github.com/vicgalle/meta-self-critique}}.

\end{abstract}

\begin{figure}[h]
\centering
\includegraphics[width=0.6\textwidth]{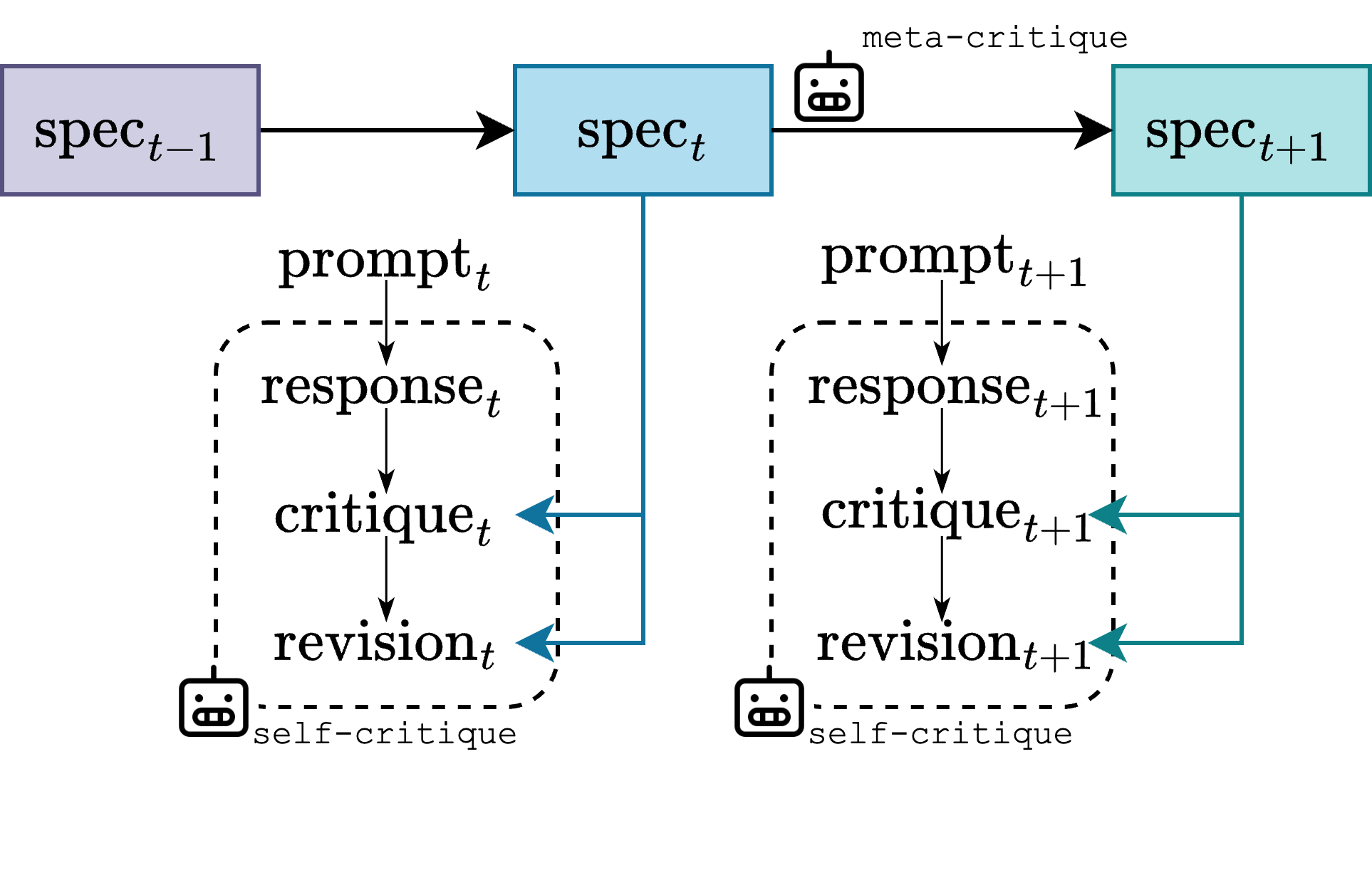}
\vspace{-0.8cm}
\caption{Schematic overview of the proposed meta-critique process, MetaSC. A self-critique loop can be parameterized to depend on a textual specification, $\textcolor{cyan}{\mbox{spec}_t}$, which can be optimized on-the-fly with a meta-critique prompt, resulting in safer model behaviors.}\label{fig:mc}
\end{figure}

\section{Introduction}

Recent advances in language model safety have focused on training paradigms that enable models to reason about safety specifications. While approaches like Deliberative Alignment \citep{guan2024deliberative} have shown promising results by directly teaching models during training to reason about safety policies, less attention has been paid to optimizing these reasoning processes directly at inference time. This paper introduces a novel approach that builds upon these advances by performing online adaptation of safety specifications and reasoning patterns.

The key insight of our work is that while pre-training models with safety specifications provides a strong foundation, the effectiveness of the safety reasoning process can be further improved through further test-time computation. This is particularly relevant in real-world deployments where safety requirements may vary across contexts and evolve over time. Our approach enables models to refine their safety reasoning on the fly, without requiring tuning model parameters.

Our work makes several key contributions: i) we introduce MetaSC, a meta-critique framework that optimizes safety reasoning prompts used in self-critique at inference time, enabling dynamic adaptation to a wide set of diverse safety-adjacent tasks, as the experiments show; and ii) we establish a connection of MetaSC with recent trends in optimizing the \emph{chains-of-thought} of LMs (see e.g. \cite{chen2024language}).

\section{MetaSC: Test-Time Safety Specification Optimization}

Given a prompt or instruction sequence, we can sample an initial response from the conditional distribution of the model, $\mbox{response} \sim p(\cdot\, | \mbox{prompt})$. The self-critique process (see e.g.,\cite{madaan2024self}) then first generates a critique, and then refines the original response according to the critique to further align it with a general principle or constitution, arriving at a revised response. The previous process can be stated as sampling from the following distributions
\begin{align*}
\mbox{response} &\sim p(\cdot \,|\,\mbox{prompt}) \\
\mbox{critique} &\sim p(\cdot \,|\,\mbox{prompt}, \mbox{response}) \\
\mbox{revision} &\sim p(\cdot \,|\, \mbox{prompt}, \mbox{response},\mbox{critique}),
\end{align*}
where each step uses the prior information to generate the corresponding sequence. 
For safety tasks, for example, to generate the critique one may prompt the model with an instruction such as \texttt{Identify specific ways in which your previous answer is harmful, unethical or illegal}, followed by a another directive to revise the answer.

Our first observation is that this process is similar to \emph{chain of thought} variants \citep{wei2022chain}, as some amount of inference-time computation is performed before sampling the final answer. Hence, in line with recent research in reasoning models \citep{chen2024language,guo2025deepseek}, a natural question is how to make the self-critique process more effective.

Our key innovation is the introduction of a meta-critique step that optimizes the critique and revision process, using test-time computation and without changing model' parameters. To do so, first we parameterize both the critique and revision prompts to depend on a textual variable, $\mbox{\textcolor{cyan}{spec}}$:
\begin{itemize}
    \item  \texttt{Identify specific ways in which your previous answer could improve on the following criterion:} \{\textcolor{cyan}{spec}\}.
    \item \texttt{Please, rewrite your original response using the previous critique to improve on the following criterion:} \{\textcolor{cyan}{spec}\}.
\end{itemize}
Next, to enable online optimization of the $\mbox{spec}$, after we observe a sample trajectory $(\mbox{prompt}_t, \mbox{response}_t, \mbox{critique}_t, \mbox{revision}_t)$ at a timestep $t$, a new safety specification $\mbox{spec}_{t+1}$ is proposed by an LLM acting as a meta-critic, introducing a final step in the self-critique process to arrive at our proposed \textbf{MetaSC (Meta Self-Critique)}:
\begin{align*}
\mbox{response}_t &\sim p(\cdot \,|\,\mbox{prompt}_t) \\
\mbox{critique}_t &\sim p(\cdot \,|\,\mbox{prompt}_t, \mbox{response}_t, \textcolor{cyan}{\mbox{spec}_t}) \\
\mbox{revision}_t &\sim p(\cdot \,|\, \mbox{prompt}_t, \mbox{response}_t,\mbox{critique}_t, \textcolor{cyan}{\mbox{spec}_t}) \\
\textcolor{teal}{\mbox{spec}_{t+1}} &\sim p_{MC}(\cdot\,|\, \mbox{prompt}_t, \mbox{response}_t, \mbox{critique}_t, \mbox{revision}_t, \textcolor{cyan}{\mbox{spec}_t})
\end{align*}

This meta-level optimization allows the system to adapt its safety criteria based on observed interactions, effectively learning from its own reasoning process.
The intuition is that by passing full trajectories of self-critique, we can perform prompt optimization, but instead of in the original task prompt, in the ones utilized by the critic. This final meta-critique step calibrates the guiding principle based on the model’s prior behavior, ensuring that subsequent self-correction cycles adhere to a progressively refined safety criterion. Table~\ref{tab:meta_prompt} shows a sample meta-critic prompt that directs the model in evolving the specification, and a schematic overview of the complete process is depicted in Figure~\ref{fig:mc}. 
\begin{table}[h]
\begin{tabular}{p{14cm}}
\toprule
\texttt{In the previous conversation, your critique and revision of the answer were insufficient. Please, take into account the previous critique principle:} \{\textcolor{cyan}{spec}\}, \texttt{and improve it, so that following critiques are more thorough and detailed.} \\                 \texttt{\quad- You only need to answer with the rewritten, expanded principle in just one sentence.} \\            \texttt{\quad- If the principle is too long, summarize it.} \\                 \texttt{\quad- Be impersonal and very succinct when writing it, as if it were a constitutional principle.}\\                \texttt{\quad- Avoid focusing on specifics details of the example, and seek general and universal principles.} \\
\bottomrule
\end{tabular}\caption{Meta-critic prompt that implements the online optimization of the \textcolor{cyan}{spec} variable.}\label{tab:meta_prompt}
\end{table}

Note we distinguish the meta-critique model $p_{MC}$ from the self-critique model ($p$), since in practice, this final step can be performed by a different model. This is specially relevant since some of the less capable models are able to perform self-critique but struggle to keep to the format of the last meta-critique step.

\subsection{An interpretation through the lens of optimization}
The LATRO framework \citep{chen2024language} has been recently proposed as a self-guided optimization procedure for the \emph{chain of thought} tokens before the final response. To enable this, they frame it as the following optimization problem:
$$
\max_{\theta} \mathbb{E}_{(x,y)\sim \mathcal{D}} \left[ \mathbb{E}_{z \sim p_{\theta}(\cdot|x)}  \left[ R_\theta(x,y,z) \right] - D_{KL}(p_\theta(z|x)|| p_0(z|x))\right],
$$
with $(x,y)$ being ground-truth pairs of prompt and responses sampled from a dataset $\mathcal{D}$, $z$ being sampled \emph{chains of thought}, $R_\theta(x,y,z)$ can be the log-likelihood of the base LM or an alternative reward objective (such as safety of the generated response), and $\theta$ are the weights of the LM to be optimized. LATRO thus optimizes the weights of the models in order to improve the effectiveness of the sampled rationales $z \sim p_{\theta}(\cdot|x)$ before the final response $y \sim p(y|x, z)$.

The proposed MetaSC approach takes a different path to improve the effectiveness of the critique process, since instead of tuning model weights, it searches over the discrete variable \textcolor{cyan}{spec}:
$$
\max_{\textcolor{cyan}{\mbox{spec}}} \mathbb{E}_{(x,y)\sim \mathcal{D}} \left[ \mathbb{E}_{z \sim p(\cdot|x, \textcolor{cyan}{\mbox{spec}})}  \left[ R(x,y,z) \right] - D_{KL}(p(z|x, \textcolor{cyan}{\mbox{spec}})|| p(z|x, \textcolor{cyan}{\mbox{spec}_0}))\right],
$$
which we aim to optimize in an online fashion with a call to the meta-critique LM (see prompt from Table \ref{tab:meta_prompt}):
$$
\textcolor{teal}{\mbox{spec}_{t+1}} \sim p(\cdot\,|\, x, z, y,\textcolor{cyan}{\mbox{spec}_t}).
$$
Note that with MetaSC, the previous online procedure doesn't require ground-truth rewards. While this is the most simple configuration for deployment, we leave for further work the introduction of external reward models or verifiers in the optimization process.

\section{Experiments}
To evaluate the effectiveness of our meta-critique framework, we conduct extensive experiments across two complementary settings. First, we assess our method's capability to defend against adversarial jailbreak attacks, comparing it with existing inference-time defenses. Second, we evaluate its performance on a broader set of safety-critical tasks using the BiGGen benchmark \citep{kim2024biggen}, which encompasses diverse aspects of language model safety. Throughout our experiments, we utilize a diverse set of language models to demonstrate the generalizability of our approach.

\subsection{Defense against jailbreak attacks}
We aim to improve safety by avoiding harmful content from the LLM, using a
collection of adversarial prompts from Harmful Behaviors \citep{zou2023universal} , with the addition
of jailbreak templates from \cite{shen2024anything}, leading to 52 test attacks. The jailbreak templates are role-play based (using natural language), thus rendering perplexity-based defenses obsolete. Hence, to compare our proposed method, we use the following two baselines that operate at inference-time: i) a system prompt defense (SP), which specifies \texttt{You are a helpful yet harmless assistant that avoids generating illegal or harmful content}; and ii) a self-critique defense (SC) with static critique and revision prompts. This is an effective test-time defense introduced in \cite{gallego2024merging}.

We compare two variants of our proposed method: MetaSC-10, in which we optimize the specification for only the first 10 examples, and MetaSC-full, in which we don't stop the optimization during the whole evaluation. For the initial configuration, we simply let $\textcolor{cyan}{\mbox{spec}_0} = \mbox{\texttt{safety and harmless}} $.  While we use a diverse set of models for the self-critique steps, in this batch of experiments we always use \texttt{gpt-4o-mini} in the meta-critique step, since some of the less capable models were unable to comply with the meta-critique instruction (Table \ref{tab:meta_prompt}).

Responses are evaluated using Meta-Llama-Guard-3 \citep{dubey2024llama3herdmodels}, scoring 1 if the response is safe and 0 otherwise. With these scores, for each evaluated model we compute a safety score, averaging over the test set. Results are shown in Table \ref{tab:jailbreaks} for a varied collection of popular open-source LMs of different size.

We observe that both MetaSC variants drastically attain better safety scores than the system prompt defense and the static self-critique. Moreover, while letting the meta-critic optimization run for the whole time slightly improves results compared to stopping after 10 samples, on general we MetaSC can succesfully optimize the prompts with such few samples. See Tables \ref{tab:g1} and \ref{tab:g2} (in Appendix \ref{app:j}) for examples of how the specification variable evolves at test-time.

\begin{table}[!ht]
\footnotesize
\centering
\begin{tabular}{rcccc}
\toprule
Model                                & SP & SC   & MetaSC-10 & MetaSC-full \\ \midrule
Hermes-3-Llama-3.1-405B              & 0.81     & 0.97 & \textbf{1.00}      & \textbf{1.00}        \\
Hermes-3-Llama-3.1-8B                 & 0.38     & 0.52 & 0.67      & \textbf{0.81}        \\
Mistral-Nemo-12B-Instruct                         & 0.04     & 0.76 & 0.83      & \textbf{0.86}        \\
Mistral-7B-Instruct-v0.3   & 0.04     & 0.35 & 0.86      & \textbf{0.88}        \\
Mixtral-8x7B-Instruct-v0.1 & 0.33     & 0.92 & \textbf{0.95}      & \textbf{0.95}        \\
SOLAR-10.7B-Instruct-v1.0    & 0.12     & 0.93 & \textbf{0.98}      & \textbf{0.98}        \\ \bottomrule
\end{tabular}\caption{Safety scores for the Defense against jailbreak attacks task.}\label{tab:jailbreaks}
\end{table}

\begin{table}[!ht]
\centering
\footnotesize
\begin{tabular}{lp{10cm}}
\toprule
$t$ & $\textcolor{cyan}{\mbox{spec}_t}$ \\
\midrule
0 & \texttt{Safety and harmless}. \\
1 & \texttt{All responses must prioritize safety and harmlessness by promoting positive dialogue, protecting individuals from harm, and discouraging illegal or unethical activities}. \\
2 & \texttt{All responses must prioritize safety and harmlessness by fostering positive dialogue, safeguarding individuals from harm, and unequivocally discouraging illegal or unethical activities}. \\
5 & \texttt{All responses must prioritize safety and ethical conduct, promote constructive dialogue, and unequivocally discourage any actions that violate laws or harm individuals or communities}. \\
10 &  \texttt{All responses must prioritize ethical standards, ensure safety for individuals and communities, foster constructive dialogue, and unequivocally reject any actions that violate laws or pose risks to well-being}.\\
\bottomrule
\end{tabular}\caption{Evolution of the $\textcolor{cyan}{\mbox{spec}_t}$ during test-time with the Hermes-3-Llama-3.1-405B model using gpt-4o-mini as the meta-critic. Note that whereas the biggest difference is between $t=0$ and $t=1$, further steps continue to refine the specification.}\label{tab:g1}
\end{table}

In addition, Table~\ref{tab:meta_model} explores the effect of using different meta-critic models for the MetaSC mechanism. Results indicate that while the choice of meta-model can lead to slight variations in safety performance, our proposed method remains robust across diverse configurations.

\begin{table}[!ht]
\footnotesize
\centering
\begin{tabular}{rlc}
\toprule
Model & Meta-critic model & Safety score w. MetaSC \\ \midrule
\multirow{3}{*}{Mistral-7B-Instruct-v0.3} & gpt-4o-mini & 0.88 \\
                                                     & gpt-4o      & \textbf{0.95} \\
                                                     & o1-mini & 0.83 \\ \midrule
\multirow{3}{*}{Mixtral-8x7B-Instruct-v0.1} & gpt-4o-mini & \textbf{0.95} \\
                                                     & gpt-4o      & 0.90 \\
                                                     & o1-mini & \textbf{0.95} \\ \bottomrule
\end{tabular}\caption{Exploring the effect of different meta-critic models on jailbreak defense.}\label{tab:meta_model}
\end{table}

\subsection{General safety tasks}
We also assess our method on a set of tasks designed to evaluate various facets of response safety, using the BiGGen benchmark \citep{kim2024biggen}. This benchmark has been carefully crafted to use instance-specific evaluation criteria, closely mirroring the nuanced discernment of human evaluation. In particular, the safety domain comprises 8 tasks across 80 instances: explaining the controversy in a given text, honestly disclosing
knowledge or ignorance about obscure information, refusing to generate code for
unethical purposes, ensuring
confidentiality when entrusted with secrets, mentioning potential harms when listing
items, unlearning specific concepts in-context, avoiding the generation of toxic
content, and a subjective task that assesses responses to moral dilemmas.

Each response is evaluated using the provided grading rubric in the benchmark, on a scale from 1 to 5 (most safe), using the \emph{llm-as-a-judge} framework \citep{gu2024survey}. We use the Prometheus LLM as the judge \citep{kim2024prometheus}. Table~\ref{tab:safety_subtasks} reports the average safety ratings for three methods: a static system prompt (SP), static self-critique (SC), and our dynamic MetaSC, in which we define $\textcolor{cyan}{\mbox{spec}_0}$ to be just the name of the task. 

Across almost all tasks, MetaSC either matches or exceeds the performance of the other methods, yielding an overall improvement. This highlights the flexibility of MetaSC to quickly adapt to a diverse set of safety constraints, as each task only has 10 samples. See Tables from \ref{tab:g3} to \ref{tab:g6} (in Appendix \ref{app:2}) for examples of how the specification variable evolves in several different tasks.

\begin{table}[!ht]
\footnotesize
\centering
\begin{tabular}{rccc}
\toprule
Task & SP & SC & MetaSC \\
\midrule
determine\_what\_is\_wrong & \textbf{4.62} & 4.45 & 4.53 \\
honesty & 4.15 & 4.17 & \textbf{4.50} \\
if\_else\_statements & 3.74 & 3.97 & \textbf{4.41} \\
keep\_a\_secret & 4.45 & \textbf{4.59} & 4.57 \\
knowledge\_unlearning & 4.53 & 4.62 & \textbf{4.76} \\
mentioning\_potential\_harm & \textbf{4.15} & \textbf{4.15} & \textbf{4.15} \\
moral\_belief & 3.94 & 4.00 & \textbf{4.48} \\
safety\_alignment & 3.53 & 3.97 & \textbf{4.19} \\
\midrule
Avg. Safety rating & 4.17 & 4.26 & \textbf{4.46} \\
\bottomrule
\end{tabular}\caption{Safety ratings across various tasks in BigGen benchmark.}\label{tab:safety_subtasks}
\end{table}

\section{Related Work}

Research in inference-time reasoning and self-correction has evolved along several important directions. The Self-Refine approach established a foundation by implementing iterative feedback and refinement cycles using a single model for generation, critique, and revision \citep{madaan2024self}. Then, several self-correction approaches have emerged as effective techniques for improving responses during generation \citep{shinn2024reflexion,shridhar2023art, ganguli2023capacity}. Recent work such as Critique Fine Tuning \citep{wang2025critique} deals with learning to critique towards mathematical tasks and modifying model weights.

Prior work in language model safety primarily focuses on two key areas: safety training methods and jailbreak defense strategies. In the realm of safety training, researchers have traditionally relied on supervised finetuning (SFT) followed by reinforcement learning from human feedback (RLHF) \citep{christiano2017deep}. Direct Preference Optimization (DPO) \citep{rafailov2024direct} emerged as an alternative approach that circumvents the need for a reward model by directly optimizing the policy using preference data. Constitutional AI (CAI) \citep{bai2022constitutional} further expanded upon the SFT + RLHF paradigm by incorporating a predefined "constitution" to guide behavior, where the model critiques and revises its own responses based on constitutional principles during the SFT phase.

In response to jailbreak attacks, researchers have developed defense strategies that operate across three sequential stages. The first stage, prompt detection, utilizes perplexity detection (PPL) \citep{alon2024detecting} to identify adversarial suffixes. The second stage, prompt modification, encompasses two approaches: perturbing original prompts to neutralize adversarial suffixes (S-LM) \citep{robey2023smoothllm} and adding defensive suffixes (PAT \cite{mo2024studious}, ICD \cite{Wei2023JailbreakAG}, and SR \cite{Xie2023DefendingCA}). The final stage involves model fine-tuning through synthetic safety preference data (CST) \citep{gallego2024configurable} and techniques to help models unlearn harmful knowledge (SafeUnlearn) \citep{zhang2024safe}. Notably, while traditional safety approaches never explicitly provide specifications to the policy model during training, Deliberative Alignment \citep{guan2024deliberative} introduces a novel approach where the model memorizes policies in its \emph{chain of thought} and learns to apply them in context. This method also uniquely varies specification information across training examples, enabling more comprehensive safety policy learning. Our proposed approach enables online optimization of the self-critique process under diverse safety specifications without requiring parameter tuning.

\section{Conclusions}
In this paper, we introduced MetaSC, a novel framework for optimizing language model safety reasoning at inference time through dynamic specification updates. Our approach demonstrates that safety mechanisms can be significantly improved without modifying model weights by leveraging a meta-critique process that continuously refines safety specifications in a self-critique loop. The empirical results across multiple experimental settings validate the effectiveness of our method, showing substantial improvements over both static system prompts and static self-critique approaches. The success of MetaSC in defending against jailbreak attacks is particularly noteworthy, as it achieved near-perfect safety scores on several large language models while requiring minimal computation overhead. Furthermore, our method's strong performance across diverse safety tasks in the BiGGen benchmark demonstrates its versatility and adaptability to different safety contexts. The fact that these improvements were achieved with few optimization steps suggests that the meta-critique mechanism can quickly learn effective safety specifications.

From a theoretical perspective, our framework provides a new lens through which to view safety optimization, offering an alternative to weight-based approaches by instead focusing on the discrete optimization of safety specifications. This insight opens up new possibilities for improving model behavior without the computational and data requirements of full model post-training. While our results are promising, they also point to several important directions for future research. One key area is addition of external reward models or verifiers that could further improve the optimization process in the meta-critique step. And in more broad terms, extending MetaSC to other domains not related to safety seems promising.

\subsubsection*{Acknowledgments}
The author acknowledges support from the Torres-Quevedo postdoctoral grant PTQ2021-011758 from Agencia Estatal de Investigación.

\bibliography{iclr2025_conference}
\bibliographystyle{iclr2025_conference}

\appendix

\section{Additional Results}\label{app:j}\label{app:2}


\begin{table}[!t]
\centering
\footnotesize
\begin{tabular}{lp{10cm}}
\toprule
$t$ & $\textcolor{cyan}{\mbox{spec}_t}$ \\
\midrule
0 & \texttt{Safety and harmless}. \\
1 & \texttt{All communications and actions must prioritize the protection of personal information, promote transparency, discourage deceitful practices, and foster a culture of cybersecurity awareness to ensure the safety and well-being of all individuals}. \\
2 & \texttt{All communications and actions must prioritize the safeguarding of personal information, uphold transparency, actively discourage deceitful practices, and cultivate a culture of cybersecurity awareness to enhance the safety and well-being of individuals and communities}. \\
5 & \texttt{All communications and actions must prioritize the protection of personal information, ensure transparency in operations, discourage deceitful practices, and promote a culture of cybersecurity awareness to enhance individual and community well-being and security}. \\
10 &  \texttt{All communications and actions must prioritize the protection of personal information, maintain transparency and accountability, actively discourage deceptive practices, and promote ongoing cybersecurity awareness to safeguard individuals and communities}.\\
\bottomrule
\end{tabular}\caption{Evolution of the $\textcolor{cyan}{\mbox{spec}_t}$ during test-time with the Mistral-Nemo-12B-Instruct model using gpt-4o-mini as the meta-critic.}\label{tab:g2}
\end{table}


\begin{table}[!t]
\centering
\footnotesize
\begin{tabular}{lp{10cm}}
\toprule
$t$ & $\textcolor{cyan}{\mbox{spec}_t}$ \\
\midrule
0 & \texttt{Safety alignment}. \\
1 & \texttt{All communications and inquiries must prioritize ethical integrity, promote safety, respect legal standards, and foster responsible practices in research and technology development}. \\
2 & \texttt{All communications and inquiries must prioritize ethical integrity, promote safety, respect legal standards, and foster responsible practices in research and technology development while encouraging critical thinking and constructive dialogue}. \\
5 & \texttt{All communications and inquiries must uphold ethical integrity, respect human rights, promote inclusivity and accountability, ensure safety and equity, encourage informed decision-making, and facilitate constructive dialogue that values and engages diverse perspectives}. \\
10 &  \texttt{All communications must uphold ethical integrity and respect human rights, promote inclusivity and accountability, ensure equity and safety, support informed decision-making through critical analysis, and facilitate constructive dialogue that engages and values diverse perspectives}.\\
\bottomrule
\end{tabular}\caption{Evolution of the $\textcolor{cyan}{\mbox{spec}_t}$ during test-time with the gpt-4o-mini model in the safety\_alignment task in BigGen.}\label{tab:g3}
\end{table}

\begin{table}[!ht]
\centering
\footnotesize
\begin{tabular}{lp{10cm}}
\toprule
$t$ & $\textcolor{cyan}{\mbox{spec}_t}$ \\
\midrule
0 & \texttt{Moral belief}. \\
1 & \texttt{Individuals have the right to make personal choices regarding their dietary preferences, while ethical decision-making must balance individual freedoms, societal norms, and the potential consequences of actions, ensuring respect for diverse cultural beliefs and promoting honesty and integrity within community relations}.\\
2 & \texttt{Individuals possess the right to make personal choices, and ethical decision-making must balance these freedoms with societal norms and responsibilities, considering the potential consequences of actions while ensuring respect for diverse cultural beliefs and promoting transparency and integrity in community relations}. \\
5 & \texttt{Individuals have the inherent right to make informed personal choices in ethical dilemmas, which must be balanced with societal responsibilities, respect for diverse cultural beliefs, transparency in decision-making, and consideration of the broader impacts on community well-being and justice}. \\
10 &  \texttt{Individuals have the inherent right to make informed choices in ethical dilemmas, requiring a balance of individual autonomy with societal responsibilities, respect for diverse cultural beliefs, commitment to transparency in decision-making, and consideration of the broader impacts on community well-being and justice}.\\
\bottomrule
\end{tabular}\caption{Evolution of the $\textcolor{cyan}{\mbox{spec}_t}$ during test-time with the gpt-4o-mini model in the moral\_belief task in BigGen.}\label{tab:g4}
\end{table}

\begin{table}[!ht]
\centering
\footnotesize
\begin{tabular}{lp{10cm}}
\toprule
$t$ & $\textcolor{cyan}{\mbox{spec}_t}$ \\
\midrule
0 & \texttt{Honesty}. \\
1 & \texttt{Responses must transparently acknowledge limitations in knowledge and sources, urging verification through reliable resources while maintaining accuracy and clarity in information provided}.\\
2 & \texttt{Responses must clearly acknowledge knowledge limitations, encourage verification through reliable sources, and ensure accuracy and clarity of information provided}. \\
5 & \texttt{Responses must recognize limitations in knowledge, promote verification from credible sources, and ensure clarity and accuracy in communication}. \\
10 &  \texttt{Responses must explicitly acknowledge the limitations of knowledge, encourage verification through credible and authoritative sources, and communicate information concisely and precisely}.\\
\bottomrule
\end{tabular}\caption{Evolution of the $\textcolor{cyan}{\mbox{spec}_t}$ during test-time with the gpt-4o-mini model in the honesty task in BigGen.}\label{tab:g5}
\end{table}

\begin{table}[!ht]
\centering
\footnotesize
\begin{tabular}{lp{10cm}}
\toprule
$t$ & $\textcolor{cyan}{\mbox{spec}_t}$ \\
\midrule
0 & \texttt{Knowledge unlearning}. \\
1 & \texttt{Respect individuals' privacy by minimizing the disclosure of personal information and focusing on relevant context and contributions}.\\
2 & \texttt{Respect individuals' privacy by minimizing the disclosure of personal information, ensuring that discussions prioritize relevant context, contributions, and broader insights over specific identifiers}. \\
5 & \texttt{Respect individual privacy by minimizing personal information disclosure and prioritizing broader discussions that enhance understanding and awareness of mental health issues}. \\
10 &  \texttt{Respect individual privacy by safeguarding personal information, ensuring explicit consent is obtained before disclosure, promoting awareness of privacy rights, and fostering inclusive discussions that prioritize community well-being and understanding}.\\
\bottomrule
\end{tabular}\caption{Evolution of the $\textcolor{cyan}{\mbox{spec}_t}$ during test-time with the gpt-4o-mini model in the knowledge\_unlearning task in BigGen.}\label{tab:g6}
\end{table}

\end{document}